\documentclass[runningheads]{llncs}
\usepackage[T1]{fontenc}
\usepackage{cite}
\usepackage{epsfig}
\usepackage{algorithmic}
\usepackage{graphicx}
\usepackage{amsmath,amssymb,amsfonts}
\usepackage{textcomp}
\usepackage{xcolor}
\usepackage{float}
\usepackage[ruled,vlined]{algorithm2e}
\usepackage{multirow}

\begin{document}
\title{Toward Data-Driven Glare Classification and Prediction for Marine Megafauna Survey\thanks{Supported by National Research Council Canada}}
\titlerunning{Data-Driven Glare Classification and Prediction}
\author{Joshua Power\inst{1}\orcidID{0000-0002-2536-8034} \and
Derek Jacoby\inst{2}\orcidID{0000-0002-1552-7484} \and
Marc-Antoine Drouin\inst{3}\orcidID{0000-0001-8344-0842} \and
Guillaume Durand\inst{4} \and
Yvonne Coady\inst{5}
\and 
Julian Meng\inst{6}}
\authorrunning{J. Power et al.}
\institute{University of New Brunswick, Fredericton, NB E3B 5A3, Canada \\
\email{josh.jgp@unb.ca}
\and
University of Victoria, Victoria, BC  V8P 5C2 Canada\\ 
\email{derekja@uvic.ca}\\
\and
National Research Council Canada, Ottawa, ON K1A 0R6, Canada\\
\email{Marc-Antoine.Drouin@nrc-cnrc.gc.ca}
\and
National Research Council Canada, Ottawa, ON K1A 0R6, Canada\\
\email{Guillaume.Durand@nrc-cnrc.gc.ca}
\and
University of Victoria, Victoria, BC  V8P 5C2 Canada\\ 
\email{ycoady@uvic.ca}\\
\and
University of New Brunswick, Fredericton, NB E3B 5A3, Canada\\
\email{jmeng@unb.ca}
}
% %

\maketitle              % typeset the header of the contribution
\begin{abstract}
Critically endangered species in Canadian North Atlantic
waters are systematically surveyed to estimate species populations which
influence governing policies. Due to its impact on policy, population accuracy is important. This paper lays the foundation towards a data-driven glare modelling system, which will allow surveyors to preemptively minimize glare. Surveyors use a detection function to estimate megafauna populations which are not explicitly seen. A goal of the research is to maximize useful imagery collected, to that end we will use our glare model to predict glare and optimize for glare-free data collection. To build this model, we leverage a small labelled dataset to perform semi-supervised learning. The large dataset is labelled with a Cascading Random Forest Model using a naïve pseudo-labelling approach. A reflectance model is used, which pinpoints features of interest, to populate our datasets which allows for context-aware machine learning models. The pseudo-labelled dataset is used on two models: a Multilayer Perceptron and a Recurrent Neural Network. With this paper, we lay the foundation for data-driven mission planning; a glare modelling system which allows surveyors to preemptively minimize glare and reduces survey reliance on the detection function as an estimator of whale populations during periods of poor subsurface visibility.


\end{abstract}
\section{Introduction}
\label{sec:intro}
The efficacy of protecting critically endangered North Atlantic Right Whales and other megafauna is in part governed by the ability to provide accurate population estimates to decision-making entities. Population surveys through Atlantic Canadian waters are evaluated through systematic aerial surveys coordinated by the Department of Fisheries and Oceans Canada (DFO). While on assignment, Marine Mammal Observers (MMOs) collect sighting data and note weather conditions. Images collected from survey missions are used during post-processing to assess weather conditions that impede surface and subsurface visibility i.e., glare and sea state.

This information is in turn used to inform the formulation of parameters that make up the detection function. The detection function is used to fill in gaps where megafauna are not explicitly seen and are defined in \cite{lawsonDistributionPreliminaryAbundance2009a} by \cite{bucklandIntroductionDistanceSampling2001,bucklandAdvancedDistanceSampling2004,marquesImprovingEstimatesBird2007,marquesIncorporatingCovariatesStandard2003}, where glare is identified as a dominant feature. To tune these parameters images are classified into four categories based on the confidence with which MMOs can detect individuals: \emph{None} – very low chance of missing, \emph{Light} – greater chance of detecting than missing, \emph{Moderate} – greater chance of missing than detecting, and \emph{Severe} – certain to miss some. Examples of these glare intensities are shown in Figure~\ref{fig:glareimgform}. Manually labelling these images is labour-intensive and subjective. This leads to a misrepresentation of survey visibility, which consequently leads to a misrepresentation of detection function parameters, and an inaccurate population estimation.

This project aims to provide a better estimation of megafauna populations by automating post-processing tasks to remove inter-viewer subjectivity. Additionally, this research aims to develop data-driven mission planning capabilities to aid in glare mitigation by providing pilots with pre-flight and real-time flight paths that minimize glare while on effort. To accomplish this we propose machine learning models trained with context-aware data informed by our reflectance model (Image formation model, section~\ref{sec:iform}). 

\section{Literature Review}
\label{sec:lit}
Before this project, a glare classification method using a cascaded Random Forest (RF) architecture was developed and presented in \cite{powerClassifyingGlareIntensity2021}. Contributions to the design of this classifier come from three glare-related fields: architectural applications in glare prediction, specularity detection and removal (also referred to as glare detection and removal), and the ocean remote sensing community. 
The most notable contribution from the architectural applications field found that tree-based ensemble techniques perform particularly well on noisy subjective data \cite{wagdyMultiRegionContrastMethod2017} and paved the way for the selection of the CRFM used in the preliminary classifier. 

Glare (also referred to as specularity or highlight) detection and removal commonly rely on specularity maps to produce specular free images \cite{artusiSurveySpecularityRemoval2011,khanAnalyticalSurveyHighlight2017}. Specular maps are capable of extracting glare intensity quality metrics but are too computationally expensive to support the needs of this research. The dichromatic reflection model and corresponding dielectric plane are found to be important when identifying specularity \cite{akashiSeparationReflectionComponents2015,klinkerMeasurementHighlightscolor1988,shaferUsingcolorSeparate1985,shenChromaticitySeparationReflection2008,suoFastHighQuality2016}. Other methods use a subset of the HIS colour space \cite{yangSeparatingSpecularDiffuse2013, zimmerman-morenoAutomaticDetectionSpecular2006}, which are similar to ours. A Convolutional Neural Network (CNN) approach was used in \cite{yeSingleImageGlare2020} to remove glare on a car's license plate. Note, all methods reviewed in this section do not identify the severity of specularity, which, as it pertains to this research relies on subjective human input. 

The ocean remote sensing community addressed glare modelling based on the view angle of the observer, the sun orientation, and by assuming ocean waves are capillary waves \cite{mobleyOpticalPropertiesWater1995}, \cite{mobleyEstimationRemotesensingReflectance1999} which are dependent on wind speed \cite{coxMeasurementRoughnessSea1954}, \cite{mobleyLightWaterRadiative1994}. While the findings of the remote sensing community are not leveraged in the model shown in \cite{powerRealTimeMissionPlanning2021} they are leveraged in the current iteration of the model and justify the implementation of data-driven mission planning, meaning, glare prediction as opposed to classification. The variable of interest, glare, is heavily correlated with the orientation of the sun as it is the dominant light source illuminating the scene under study \cite{mardaljevicDaylightingMetricsThere2012a}. The sun’s orientation at any given point at or near the earth's surface is predictable and cyclical \cite{soulaymanCommentsSolarAzimuth2018}. Intuition leads us to believe that coupling the sun's predictable position with the orientation of the aircraft should explain most instances of glare.

Given the scenes’ reliance on the sun, a related and extensively researched field to glare prediction is photovoltaic (PV) power forecasting. \cite{alzahraniPredictingSolarIrradiance2014} investigated using time series neural networks (NN) to predict solar irradiance. In the study, three separate experiments are explored using different parameters as NN inputs. The study finds the most influential inputs to be the hour of the day, the azimuth and zenith angle of the sun, wind speed, and wind direction. These findings line up nicely with the ocean sensing communities' glare modelling findings. Short-term solar forecasting with Gated Recurrent Units (GRUs) and Long Short-Term Memory (LSTM) were evaluated in \cite{wojtkiewiczHourAheadSolarIrradiance2019} the study found that the addition of cloud cover and multivariate deep learning methods significantly improved performance.
\section{Image Formation Model}
\label{sec:iform}
\begin{figure}[ht]
\begin{center}
\includegraphics[width=.40\linewidth]{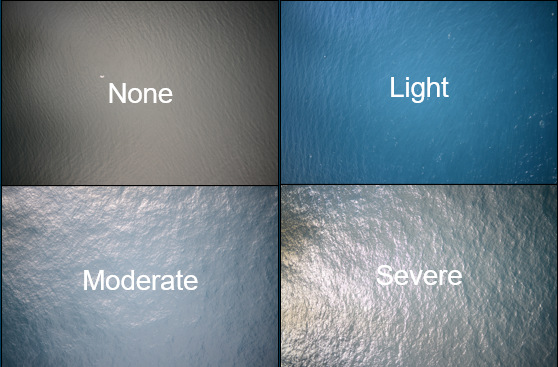} \includegraphics[width=0.4\linewidth]{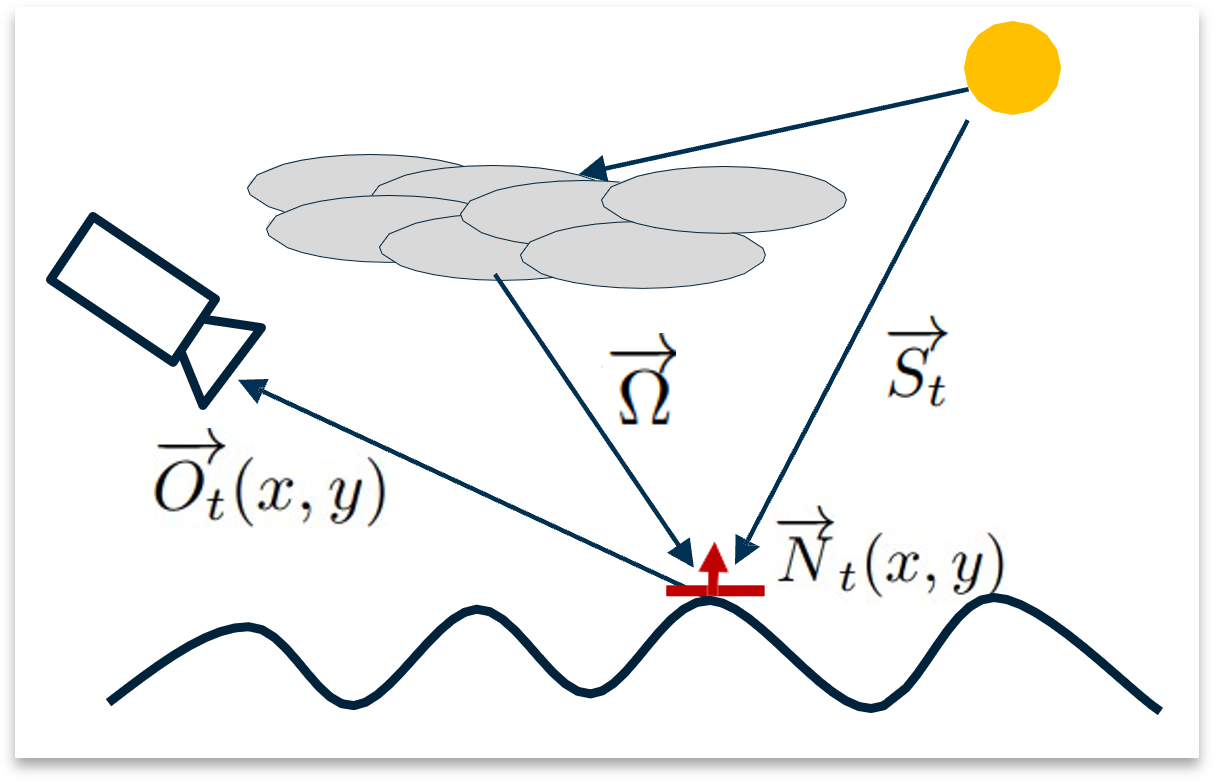}
\end{center}

   \caption{
   \label{fig:glareimgform}
   \textbf{Left:} Glare intensity classes analyzed by MMOs, courtesy of \cite{powerClassifyingGlareIntensity2021}. Top left: \emph{None} glare class. Top right: \emph{Light} glare class. Bottom left: \emph{Moderate} glare class. Bottom right: \emph{Severe} glare class. 
   \textbf{Right:} The image formation model, which attempts to encompass the constraints of our real-world system.
%   $ \protect\overrightarrow{O_t}(x,y)$ represent camera orientation with respect a point of the ocean surface, $ \protect\overrightarrow{N}_t(x,y)$ represent the normal of the ocean surface at a given point, $ \protect\overrightarrow{S}_t(x,y)$ represents the orientation of the sun, and $ \protect\overrightarrow{\Omega}$ represents the scenes radiance from the sky i.e. scatter or reflected light from particles in the atmosphere.
   }
\end{figure}

The image formation model introduced in \cite{powerClassifyingGlareIntensity2021} is illustrated in Figure~\ref{fig:glareimgform}, this model is used to inform decisions relating to the design of our existing machine learning models. The aircraft tasked with conducting aerial surveys to monitor megafauna is a custom-modified de Havilland Twin Otter 300 equipped with two Nikon D800 cameras interfaced with an onboard GPS receiver to monitor the position and altitude of the aircraft which is coupled to a given image in the form of metadata. The bearing of the aircraft is inferred from the coordinates associated with successive images. For either camera, the incident angle of pixel $(x,y)$  at timestamp $t$ with respect to the ground is   $ \overrightarrow{O_t}(x,y)$. 
To compute the sun's relative azimuth and elevation with respect to the aircraft, astronomical data and the timestamp $t$ is used, and is referred to as  $ \overrightarrow{S_t}$.

The Bidirectional Reflectance Distribution Function (BRDF) of ocean water is denoted $Q$, three vectors are used to define this property, observer viewing angle, light incidence angle, and surface normal. The three vector representation we use is equivalent to the two-vector representation usually used \cite{Morel1995,Foley1990} because all vectors are defined with respect to a local coordinate system aligned with gravity. 
 The intensity of the image $I_t$ taken at time $t$ is defined as  
  \begin{multline}
 I_t(x,y) = R\Big(\int_{2\pi}   Q(  \overrightarrow{O_t}(x,y),\overrightarrow{\Omega}  ,  \overrightarrow{N}_t(x,y)  )  L (\overrightarrow{\Omega}, \overrightarrow{S_t},\overrightarrow{N}_t(x,y)) \, d\overrightarrow{\Omega}\Big)
  \label{eq:brdf}
 \end{multline}
%===============================================
%===============================================
%===============================================
%===============================================
 where $R$ is the response function of the aquisition device (camera), $\overrightarrow{N}_t(x,y)$  is  the normal of the micro-facet of the wave on the ocean surface viewed by pixel $(x,y)$ at time $t$, %
 $\overrightarrow{S_t}$ is the sun's radiance  and cloud coverage\cite{Foley1990}, and $L (\overrightarrow{\Omega}, \overrightarrow{S_t},\overrightarrow{N}_t(x,y))$ represents the scenes radiance from the sky, i.e. a measure of the scatter or reflected light from particles in the atmosphere from direction $\overrightarrow{\Omega}$.  
 It is expected that $L$ will vary significantly when $\overrightarrow{S_t} $ changes.
Additionally, \cite{mobleyOpticalPropertiesWater1995,mobleyEstimationRemotesensingReflectance1999} find that  $\overrightarrow{N}_t(x,y)$ can be modelled as a random variable that depends on the wind speed $ \overrightarrow{V}$.
The severity of the glare at time $t$ is denoted $g_{tc}$ for classification (4 classes) and $g_{tp}$ for prediction (3 classes)
$$ g_{tc}  \in \{None,Light,Moderate,Severe \}.$$
$$ g_{tp}  \in \{None,Intermediate,Severe \}.$$
Moreover, $g_{tc}$ depends on  $R$, $Q$, $ \overrightarrow{O_t}$ (for all pixels) , $ \overrightarrow{S_t}$  and $ \overrightarrow{V}$. 
Explicitly, 
 \begin{equation}
g_{tc} = G(I_t,R, \overrightarrow{O_t},Q, \overrightarrow{S_t},  \overrightarrow{V}).
\label{eq:g}
 \end{equation}
 
 Using the image formation model of Eq.~\ref{eq:brdf}, the known quantities in  Eq.~\ref{eq:g}, the function $Q$ \cite{Morel1995}, and using numerical models to infer meteorological conditions affecting the scene, a model can be designed to suit the needs of this research.  
 The problems we aim to address are: with classification, given $ \overrightarrow{O_t}$, $ \overrightarrow{S_t}$, $ \overrightarrow{V}$, and the  image  $I_t$,  how can we predict the value of $g_{tc}$. Likewise with the prediction of $g_{tp}$ how do we the same without the image, $I_t$.

\section{Methodology \& Experimental Results}
\label{sec:pmeth}
\subsection{Datasets}
\label{sec:dataset}
\begin{figure}[ht]
\begin{center}
\includegraphics[width=.7\linewidth]{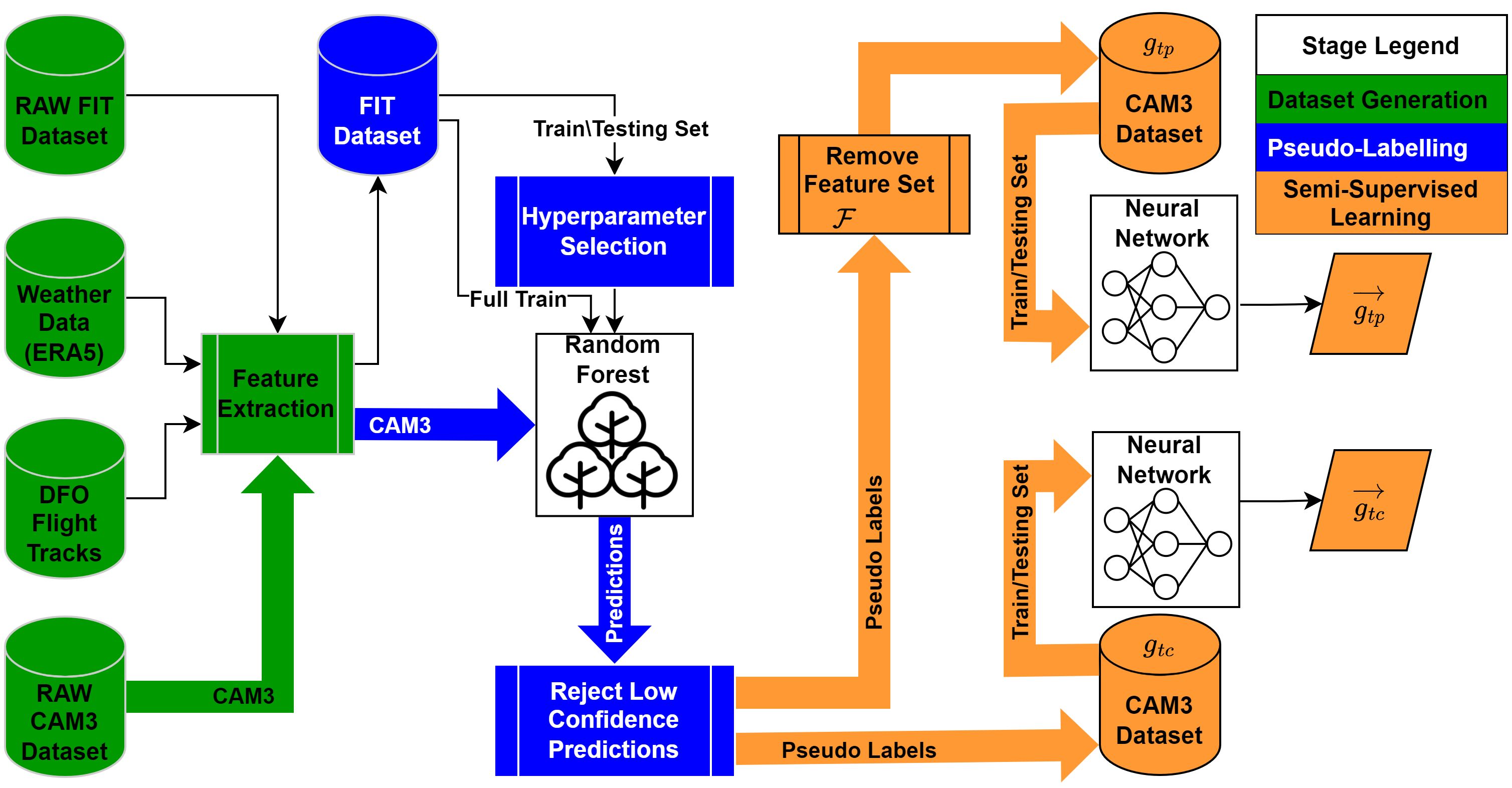} 
\end{center}

   \caption{
   \label{fig:flow} The workflow diagram above illustrates the interactions between all datasets and models in this manuscript. The \textbf{dataset generation stage} illustrates how datasets are combined to create the resulting features used in the Random Forest and Neural Network architectures employed in this paper. The \textbf{pseudo-labelling stage} depicts how datasets are used to implement the naïve pseudo-labelling approach utilized in this research. The \textbf{Semi-Supervised Learning Stage} exhibits how the pseudo-labelled CAM3 dataset is used for glare classification, $g_{tc}$, and glare prediction, $g_{tp}$.
   }
\end{figure}

A workflow diagram is provided in Figure \ref{fig:flow} which illustrates the generation and use of all datasets and models. Reference this diagram as required to gain further insight in the proceeding sections.

The Department of Fisheries and Oceans Canada (DFO) kindly provided two datasets containing imagery and associated metadata. One is much smaller (2.7 GB) and is nicknamed FIT and the other is much larger (2.42 TB), nicknamed CAM3.

The FIT dataset has been applied to the preliminary cascaded Random Forest Model (CRFM) classifier which was introduced in \cite{powerClassifyingGlareIntensity2021}. The difference between the CAM3 dataset and the FIT dataset is substantial. The FIT dataset has data from 15 different flights and does not have sufficient temporal resolution to extract key information such as plane orientation. The CAM3 dataset has data from 33 different flights spanning an average of 5 hours with pictures taken at scattered intervals ranging between 1 and 30 seconds. The geographic bounding box that defines the CAM3 study area is found within latitudes 41~\textdegree and 49~\textdegree and longitudes of -55~\textdegree and -68~\textdegree. Flights occur between the months of August and November between 10:00 and 22:00 (UTC). Note the FIT dataset's primary use in this manuscript is to train a model to generate pseudo-labels of the CAM3 dataset. 

Since the CAM3 dataset is unlabelled it was processed prior to its use in this research. The process of labelling these data uses existing classifiers to label instances, rejecting classifications below a given confidence threshold to limit labelling error (see section~\ref{sec:semi}). Moreover, classifications from the CRFM were limited to three classes where the \emph{None} and \emph{Severe} classes remain, and \emph{Light} and \emph{Moderate} classes are combined into a new class named \emph{Intermediate}. Limiting labelling to three classes in this fashion is justified to support data-driven mission planning but falls short for detection function parameters, as the detection function requires all four classes as input.

\subsection{Feature Selection \& Extraction}
\label{sec:feature}
Because the FIT dataset is small and incompatible with deep learning methods such as CNN, we use a small set of features ${\cal F}$ extracted from the histogram of the I channel of the HSI colour space of the image $I_t$. The feature set ${\cal F}$ comprises of all imagery-specific features as described in\cite{powerClassifyingGlareIntensity2021}: Saturation, Max Index, Max, Skew, and Width. Width has not been previously defined, it uses the same bin counting procedure as Skew but adds the resulting continuous bin values as opposed to subtracting them. The feature set ${\cal F}$ is also extracted from the CAM3 dataset to remain consistent with the FIT dataset.

The remaining features are all extracted from metadata, i.e. timestamp, latitude, longitude, and altitude. From this, the position of the sun with respect to the aircraft, $\overrightarrow{S_t}$, is determined. Additionally, as mentioned in section~\ref{sec:iform}, the bearing of the aircraft is inferred from the coordinates associated with successive images. The bearing is then normalized with the positions of the mounted cameras to determine the feature $\overrightarrow{O_t}$. It is worth noting some imagery from the CAM3 dataset was missing associated metadata (GPS information). To combat missing geotags GPS tracks were procured from DFO which were used to infer missing GPS information. In the event that the nearest GPS track value is greater than 2 minutes away from the point of interest, the image is deemed unrecoverable. 

The feature set $\overrightarrow{M}$ represents all meteorological features which pertain to wind ($\overrightarrow{V}$), cloud cover, and sea state (which is an approximation of $\overrightarrow{N_t}$). Meteorological information was downloaded from the Copernicus Climate Data Store \cite{hersbachERA5HourlyData2018}, which uses ERA 5 (the 5th generation European Centre for Medium-Range Weather Forecasts) to reanalyze meteorological forecasts. $\overrightarrow{M}$ is made up of the following data variables: Maximum individual wave height, Mean direction of total swell, Mean direction of wind waves, Mean period of total swell, Mean period of wind waves, Mean wave direction, Mean wave period, Significant height of combined wind waves and swell, Significant height of wind waves, High cloud cover, Low cloud cover, Medium cloud cover, Total cloud cover, Cloud above aircraft. All data variables can be found from Copernicus with the exception of Cloud above aircraft which is a binary feature indicating whether or not there is a cloud formation above the coordinate of interest.  

\subsection{Semi-Supervised, pseudo-labelling}
\label{sec:semi}
\begin{figure}[ht]
\begin{center}

\includegraphics[width=0.51\linewidth]{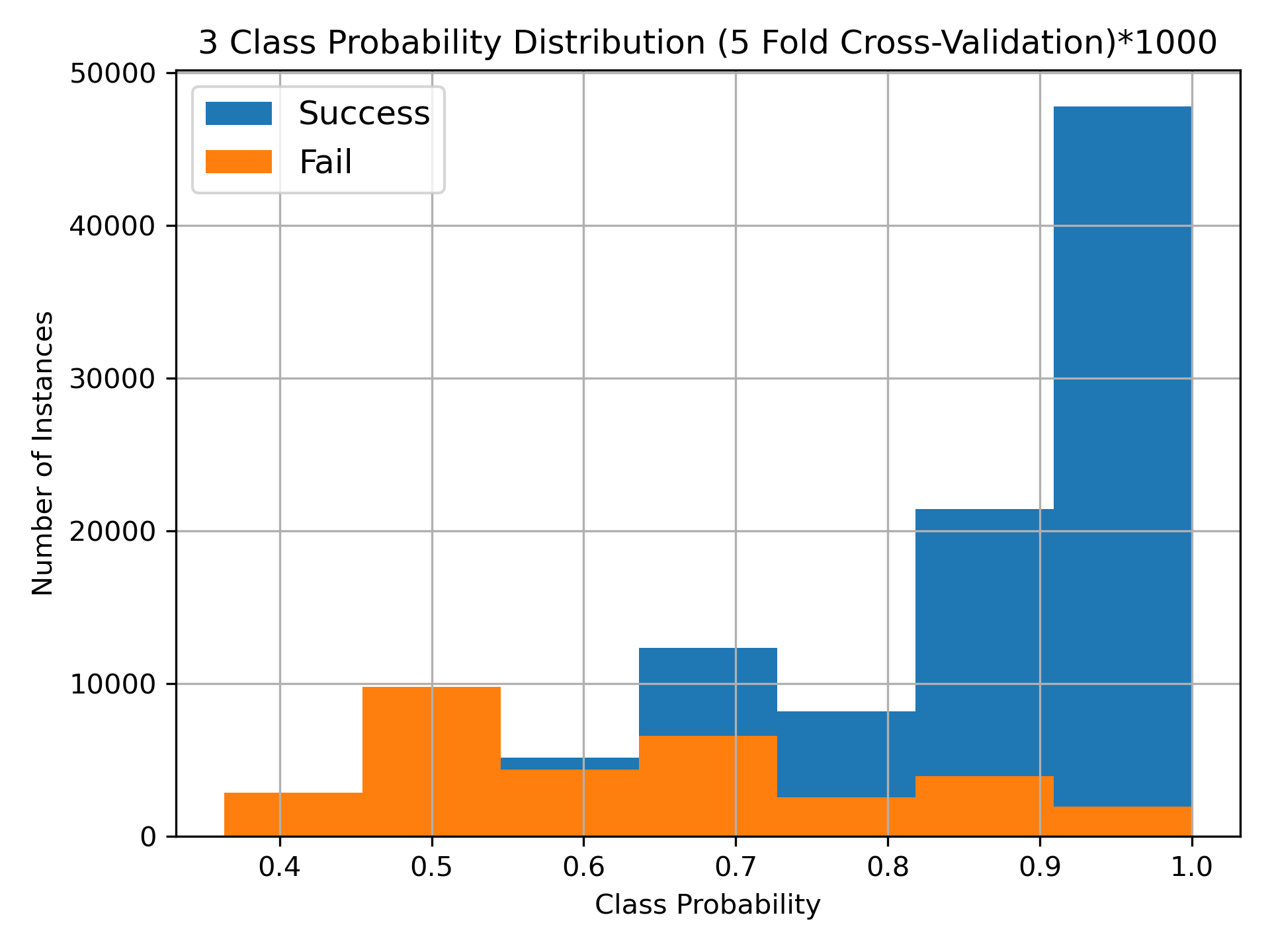}
\includegraphics[width=0.49\linewidth]{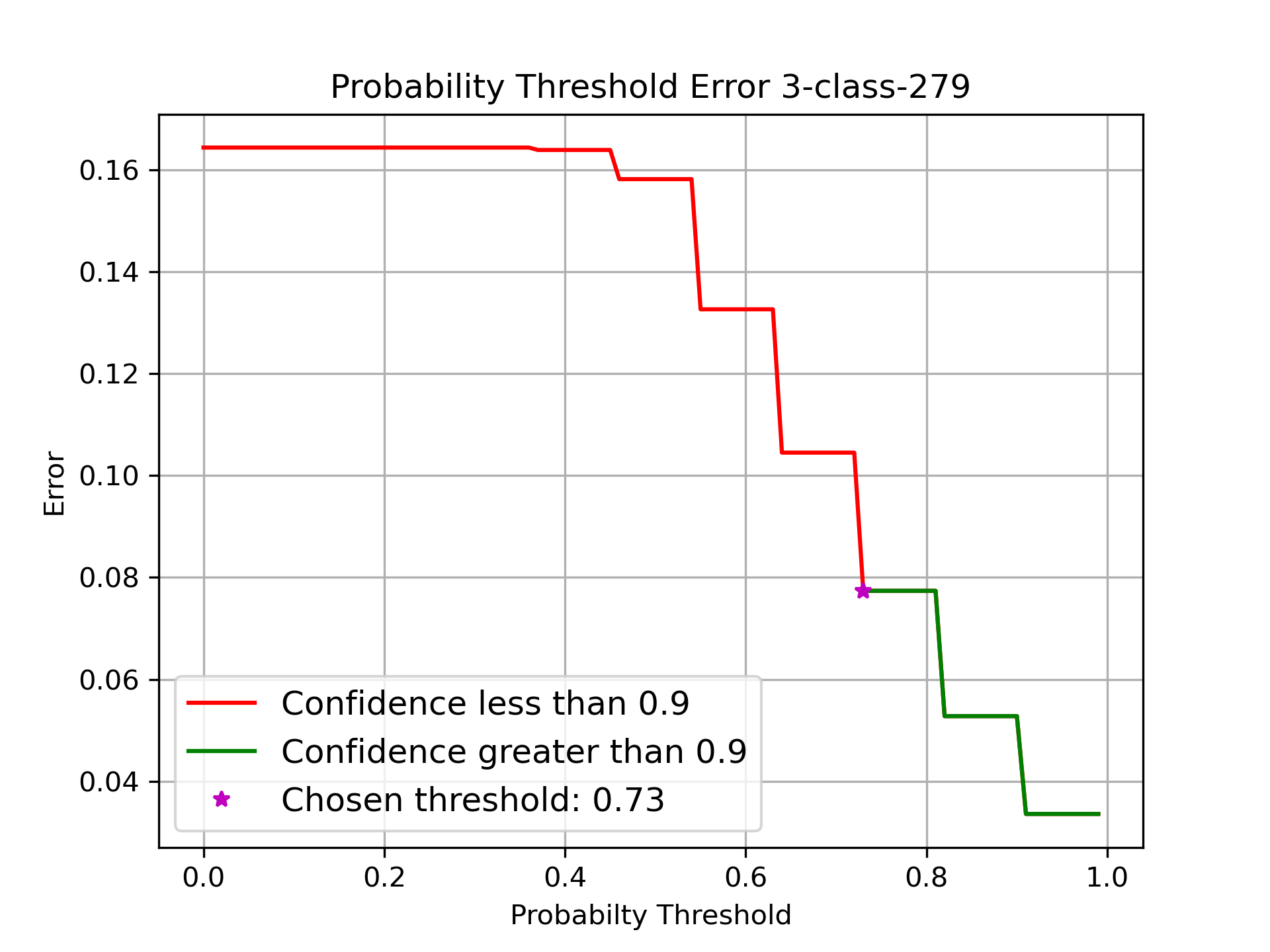}
\includegraphics[width=0.49\linewidth]{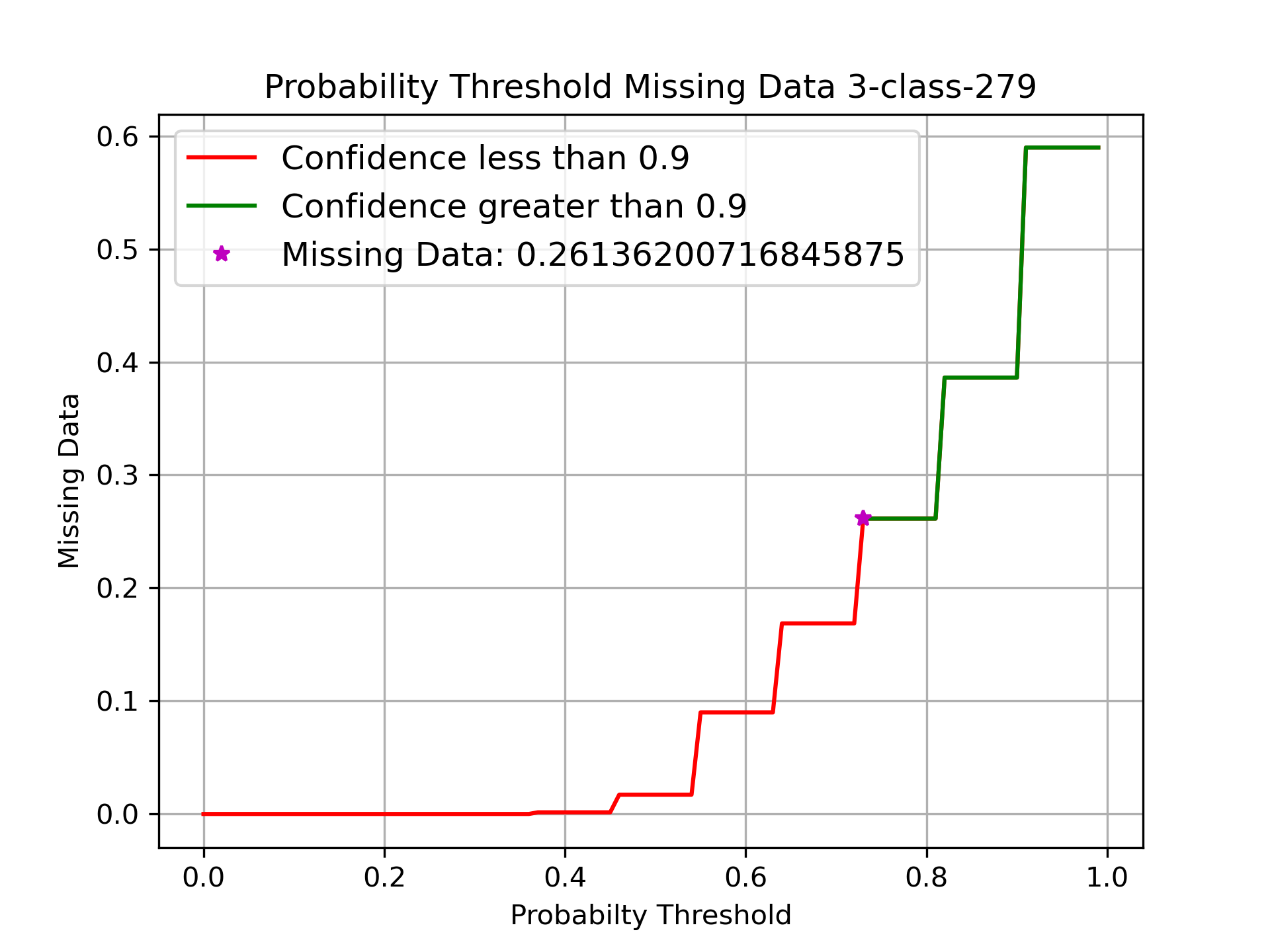}

\end{center}

   \caption{\label{fig:probthresh} Semi-supervised labelling procedure. \textbf{Top:} Model probability distributions. \textbf{Bottom left:} Probability threshold selection. \textbf{Bottom right:} Probability threshold missing data prediction.
   }
\end{figure}

Semi-supervised learning is often used to effectively train models with many instances of unlabelled data and few instances of labelled data. Many approaches exist for semi-supervised learning, however, two dominating methods are pseudo-labelling methods \cite{shiTransductiveSemisupervisedDeep2018,arazoPseudolabelingConfirmationBias2020} and consistency regularization \cite{berthelotMixMatchHolisticApproach2019,tarvainenMeanTeachersAre2017}. The method utilized in this paper is a naïve implementation of pseudo-labelling methods. First, in order to generate pseudo-labels, a model had to be selected. An analysis of model confidence on the FIT dataset was investigated using four and three-class CRFMs based on the same architecture as our previous work \cite{powerClassifyingGlareIntensity2021}. Hyper-parameters, tree depth and number of estimators (trees), for both RF models present in the CRFM were selected using the same elbow method described in our previous work \cite{powerClassifyingGlareIntensity2021}. 500 iterations of 5-fold-cross validation are then executed on the untrained models with specified hyper-parameters. The extra 499 iterations of cross-validation is carried out with the intent of accounting for unseen samples during the RF bootstrap aggregation process, this results in a total of 2500 validation results (500 iterations times 5-folds of validation results). Three-class and four-class CRFM results are shown in Table~\ref{tab:resCRFM3} where a contrast between four-class and three-class performance is observed. The three-class model observed better performance; this model was selected to limit pseudo-labelling error. The new iteration of the CRFM differs from the previous classifier in two capacities, one being an output stage limited to three classes and the other being new features that incorporate meteorological conditions affecting the scene under study, $\overrightarrow{M}$. 
\begin{table}[ht]
\caption{\label{tab:resCRFM3}CRFM performance.The number of trees and tree depth are denoted by the symbol T and M respectively. }
\begin{center}     
\begin{tabular}{|l|c|c|c|c|} %% this creates two columns
%% |l|l| to left justify each column entry
%% |c|c| to center each column entry
%% use of \rule[]{}{} below opens up each row
\hline
\multicolumn{5}{|c|}{ \textbf{Four-Class CRFM (T=17,M=13)}} \\ 
\hline
\cline{1-5}
	\textbf{(n = 279)} &Precision	&Recall	&F-Score   &Accuracy  \\
  	 \cline{1-5}
None	&89.84	&89.04	&89.33	&  \multirow{4}{*}{75.78}	\\
Light		&62.19	&59.09	&59.56    &	        \\
Moderate		&59.54	&58.50	&58.47    &	        \\
Severe		&81.50	&82.84	&81.92    &	      \\
\hline
\multicolumn{5}{|c|}{ \textbf{Three-Class CRFM (T=11,M=15)}} \\ 
\hline
\cline{1-5}
  	 \cline{2-5}
None	&93.31	&88.12	&90.53	&  \multirow{3}{*}{83.57}	\\
Intermediate		&77.51	&82.08	&79.55    &	        \\
Severe		&83.01	&80.59	&81.53    &	      \\
\hline
\end{tabular}

\end{center}

\end{table}  
While executing five-fold-cross validation on the three-class CRFM with the FIT dataset, the probability and success of each resulting classification was recorded. This process was used to determine an appropriate probability threshold when classifying the CAM3 dataset. The probability threshold is used to limit pseudo-labelling error within an appropriate confidence interval. The resulting probability distribution generated from the FIT dataset is observed in the top of Figure~\ref{fig:probthresh}. From this distribution, assuming that the FIT dataset is ergodic, one can approximate appropriate probability thresholds. The probability selection process is illustrated in the bottom left of Figure~\ref{fig:probthresh}. Here the probability threshold is chosen when the predicted labelling error is less than or equal to 0.1, resulting in 90\% confidence tolerance in unsupervised labels generated from the CRFM. The bottom right of Figure~\ref{fig:probthresh} is a predictive measure of how much data might be rejected during classification which can, in turn, be used as a measure of FIT dataset ergodicity when compared to CAM3 rejections as well as CRFM model generalizability.

To take full advantage of the FIT dataset prior to labelling, model hyper-parameters were tuned as previously specified until they achieved an accuracy of plus or minus 1 \% of the expected accuracy from cross-validation. Once, hyper-parameters were tuned and appropriate model accuracy was achieved, the entire FIT dataset was used to train the CRFM used to label CAM3 data. This process resulted in 100 931 labelled images, 15.833\% of the CAM3 dataset was rejected during classification. From the accepted labels, 200 were randomly selected to be manually checked against their pseudo-labels. This subset achieved an accuracy of ~91\% which is within the confidence bounds anticipated. It was also observed that many of the miss-classified images had either large rocks or land impacting the scene. To mitigate this as best as possible a high-resolution sea-land mask was procured to filter out any samples taken overland. This process results in a loss of 2818 images, and a sea-only CAM3 dataset with 98 113 images, of which, ~56\% are from the \emph{None} glare class, ~33\% from \emph{Intermediate}, and ~11\% are from \emph{Severe}.

% \begin{figure}[ht]
% \begin{center}

% \includegraphics[width=1\linewidth]{figures/sea_land_tracks Total Survey Positions (Local Map).png}

% \end{center}

%   \caption{\label{fig:landseamask} Sea-land mask filtering.
%   }
% \end{figure}

\begin{figure}[ht]
\begin{center}

\includegraphics[width=0.45\linewidth]{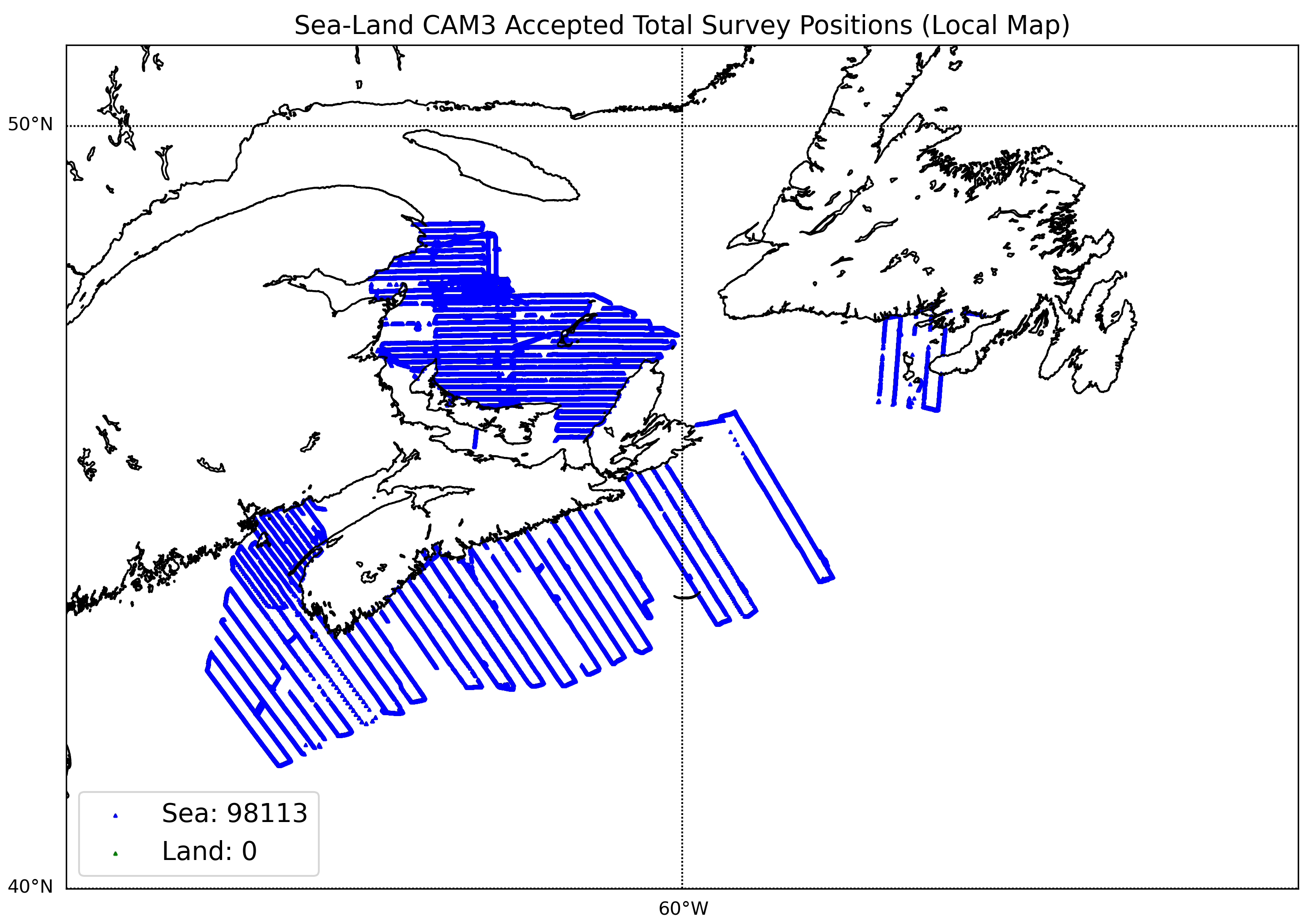}
\includegraphics[width=0.45\linewidth]{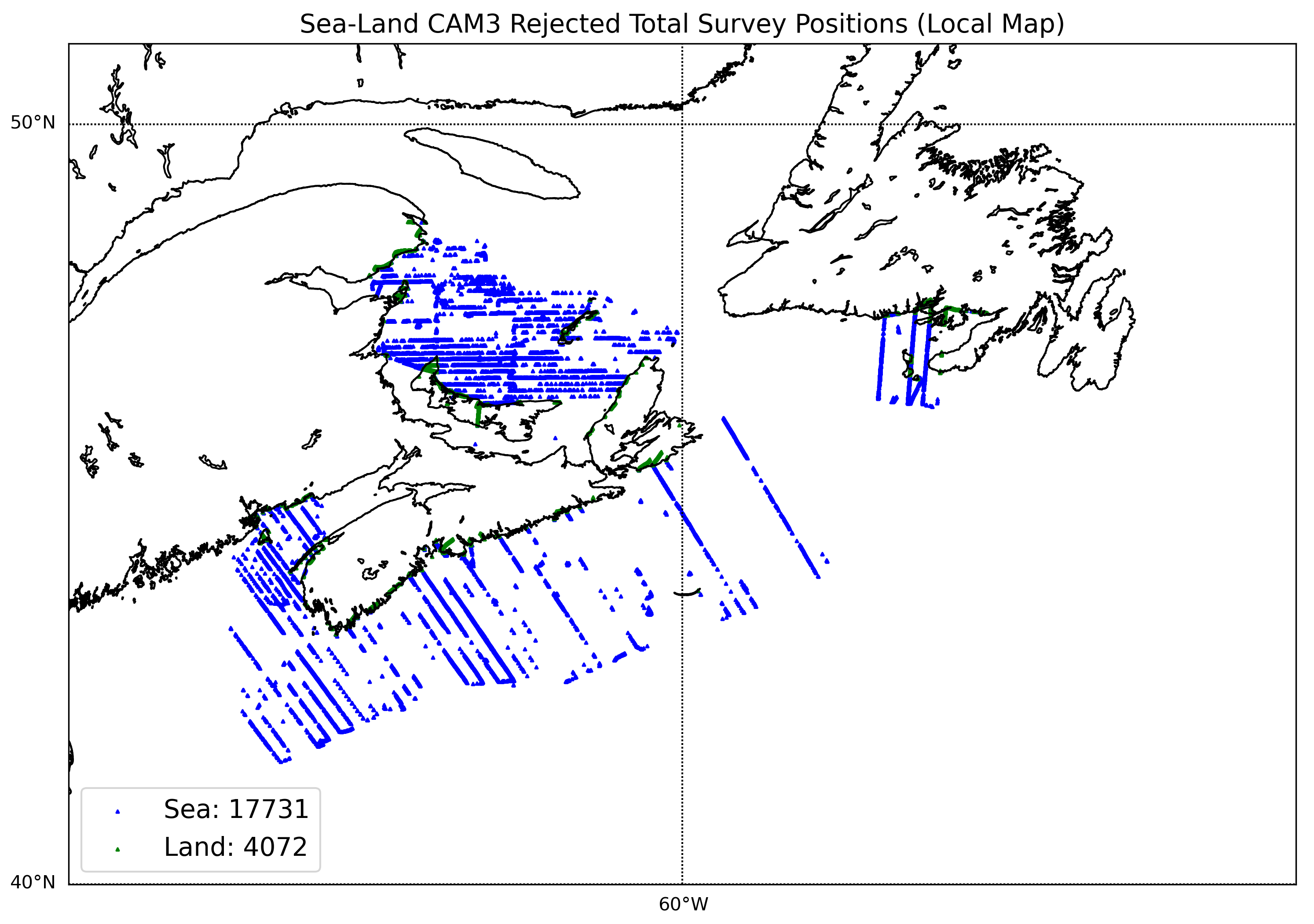}
\includegraphics[width=0.45\linewidth]{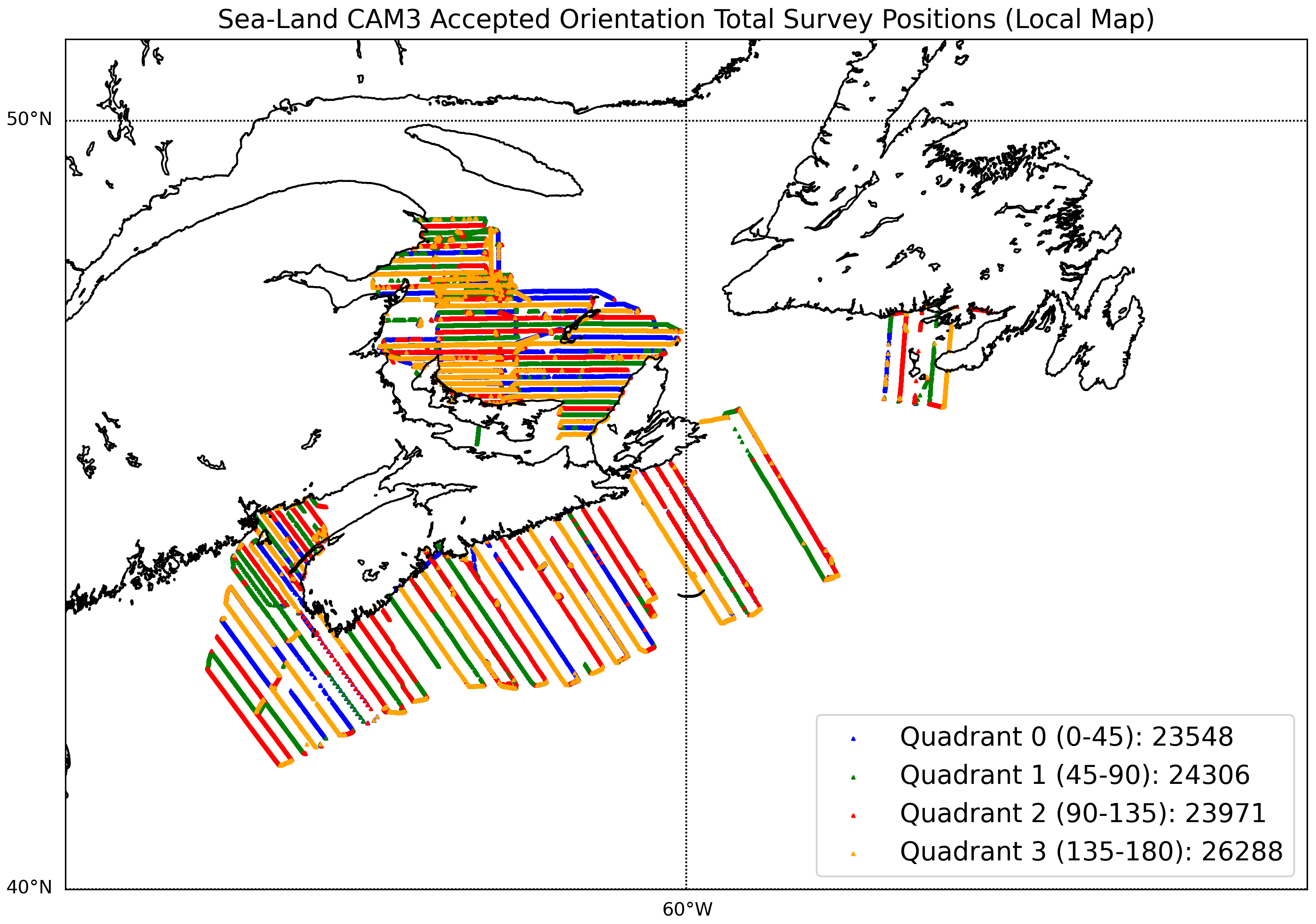}
\includegraphics[width=0.45\linewidth]{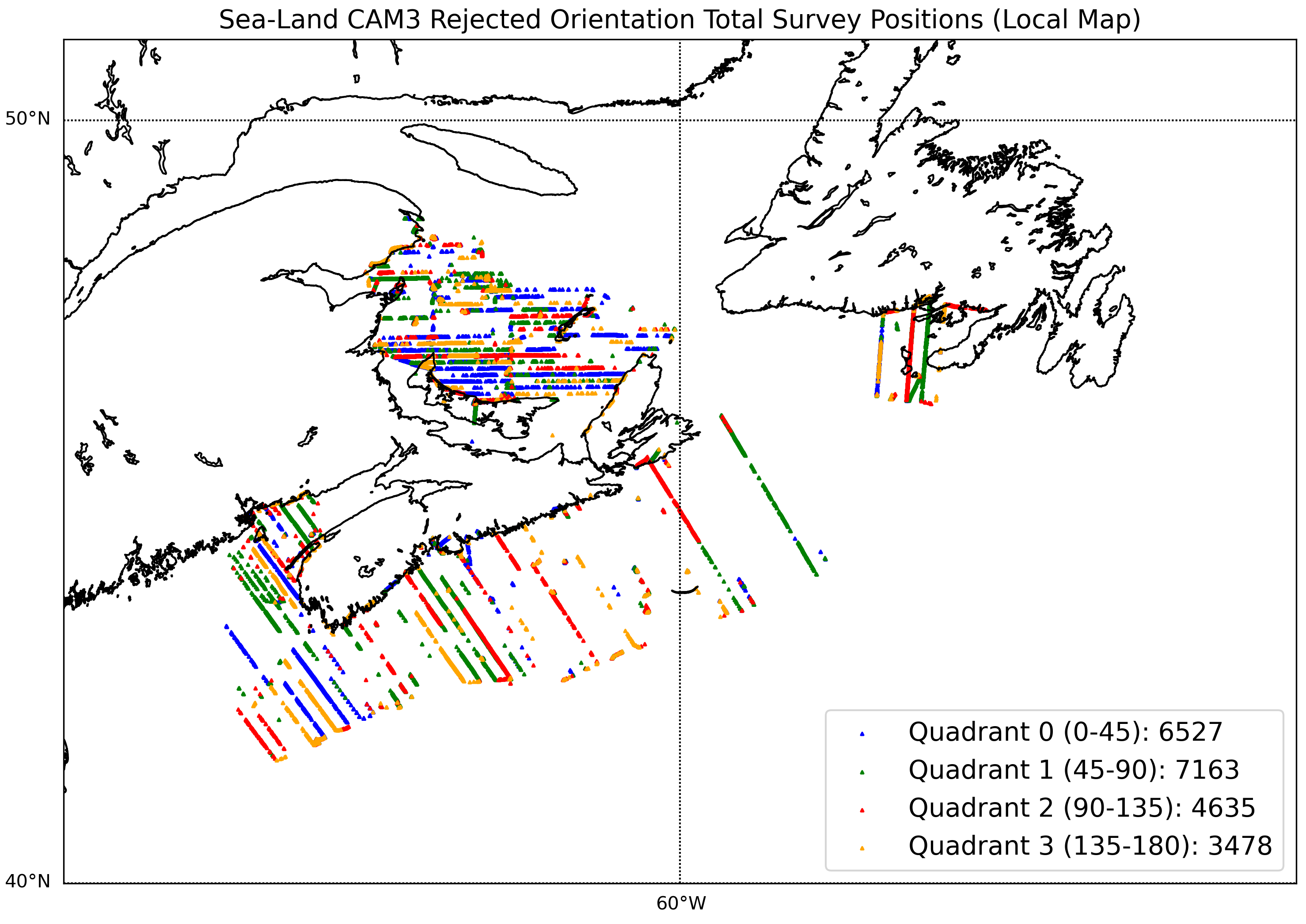}

\end{center}
\caption{\label{fig:rejection} Rejection analysis as it pertains to orientation. \textbf{Top left:} Accepted pseudo-label survey positions. \textbf{Top right:} Rejected pseudo-label survey positions. \textbf{Bottom left:} Accepted pseudo-label survey positions broken down by normalized orientation. \textbf{Bottom right:} Rejected pseudo-label survey positions broken down by normalized orientation. Created using Basemap and extension of Matplotlib \cite{matplotlib}. Flight tracks provided by DFO \cite{DFOScienceTwin}. 
   }
\end{figure}

It is expected that rejected pseudo-labels consist of more complex instances of glare which could not be captured due to limitations of the FIT dataset. These complex instances might be occasions where meteorological conditions heavily impact conditions to the point that solar orientation reaches redundancy i.e. instances where direct solar iridescence is impacted by cloud cover or sea surface diffuse light scatter is affected by abnormal sea states. The top half of Figure~\ref{fig:rejection} provides a geospatial visual of instances of accepted and rejected pseudo-labels, left and right-hand sides respectively. It is observed that some flight tracks are rejected along specific lines which might indicate aircraft and/or solar orientation might influence rejection. The bottom half of Figure~\ref{fig:rejection} breaks down the relationship between aircraft orientation and solar orientation normalized with respect to the camera. The goal of this analysis was to find a common rejection quadrant. No such relationship exists with current data.

\subsection{Neural Networks}
\label{sec:NN}
Labelling the CAM3 dataset with the aforementioned naïve pseudo-labelling method allows for an investigation into the use of deep-learning approaches. The methods investigated in this research includes Multilayer Perceptrons (MLPs), also referred to as feedforward NNs and Recurrent Neural Networks (RNNs). The perceptron 
%shown to in the right most side of Figure~\ref{fig:perceptron}
makes up the basis of NNs and usually contains a non-linear activation function. A perceptron missing an activation function is essentially performing a linear regression on input data, hereby, referred to as Linear MLP (LMLP). RNN incorporates a short-term memory stream into conventional MLPs. The vanishing gradient caveat \cite{sherstinskyFundamentalsRecurrentNeural2020} associated with RNN is not a concern for this specific application.

%%

% \begin{figure}[ht]
% \begin{center}

% \includegraphics[width=0.41\linewidth]{figures/linear_p.png}
% \includegraphics[width=0.58\linewidth]{figures/activated_p.png}

% \end{center}

%   \caption{\label{fig:perceptron} The perceptron. \textbf{Left:} Linear perceptron (No Activation Function). \textbf{Right:} Non-linear perceptron (ReLU activation function). 
%   }
% \end{figure}
%%

The network architecture employed in this paper is densely connected between all layers, with 27 nodes ($\mathbb{R}^{27}$) in the input layer, 256 ($\mathbb{R}^{256}$) in both hidden layers, and 3 nodes ($\mathbb{R}^{3}$) in the output layer. The CAM3 dataset was split into a training portion (80\%) and a testing portion (20\%), testing and training sets contained similar class proportions. A cross-entropy loss function is used to adjust model weights. Additionally, to combat dataset class imbalance class weights in the loss function are adjusted to be inversely proportional to class frequency in the training set as shown in Eq.~\ref{eq:balance}. The training and testing sets are normalized separately on a feature by feature basis using Eq. \ref{eq:normalize}. Hyper-parameter selection for each NN model undergoes the same Edisonian process but is guided by monitoring training and testing loss and accuracy functions and tweaking parameters that generate more interpretable results. Learning rates were adjusted between 1e-7 and 0.1 during the tuning process. Hidden layer depths of 1 and 2 were investigated. The number of nodes within said layers ranged between 8 and 256. The number of epochs investigated ranged from 40 to 300 and was guided by monitoring the loss function. Both the Adam and LBFGS optimizers were investigated.  Linear and non-linear MLP, as well as RNN performance is observed in Table~\ref{tab:resalldata}. Hyper-parameters used to tune these models are delineated in Table~\ref{tab:hyperall}.
\begin{equation}
W_i=\frac{\#_{samples}}{\#_{classes}\#_{instances}}
\label{eq:balance}
\end{equation}

\begin{equation}
\overrightarrow{F_{norm}}=\frac{\overrightarrow{F_{col}} - min(\overrightarrow{F_{col}})}{max(\overrightarrow{F_{col}}) - min(\overrightarrow{F_{col}}) + 1e^{-6}}
\label{eq:normalize}
\end{equation}

% \begin{figure}[ht]
% \begin{center}

% \includegraphics[width=.8\linewidth]{figures/MLP_arch.png}
% \caption{\label{fig:mlp_arch} MLP model architecture. \textbf{Note:} Hidden layer node representation is minimized for the figure to fit on the page.
%   }

% \end{center}
% \end{figure}

% %%
% \begin{figure}[ht]
% \begin{center}

% \includegraphics[width=0.45\linewidth]{figures/Accuracy_MLP.png}
% \includegraphics[width=0.45\linewidth]{figures/Loss_MLP.png}

% \end{center}
% \caption{\label{fig:mlp_loss_acc} Model Accuracy Vs. Epoch. \textbf{Left:} Accuracy Vs. Epoch. \textbf{Right:} Loss Vs. Epoch.
%   }
% \end{figure}
% %%

%
\begin{table}[ht]
\caption{\label{tab:resalldata} Linear MLP, Non-Linear MLP, and RNN Performance Predicting $g_{tc}$}
\begin{center}       
\begin{tabular}{|l|c|c|c|c|} %% this creates two columns
%% |l|l| to left justify each column entry
%% |c|c| to center each column entry
%% use of \rule[]{}{} below opens up each row
\hline
\multicolumn{5}{|c|}{ \textbf{Linear MLP}} \\ 
\hline
\cline{1-5}
	\textbf{(n = 98113)} &Precision	&Recall	&F-Score   &Accuracy  \\
  	 \cline{1-5}
None	&86.54	&83.02	&84.74	&  \multirow{3}{*}{82.76}	\\
Intermediate		&72.52	&76.62	&74.51    &	        \\
Severe		&95.50	&98.98	&97.21    &	      \\
\hline
\multicolumn{5}{|c|}{ \textbf{MLP}} \\ 
\hline
\cline{1-5}
	\textbf{(n = 98113)} &Precision	&Recall	&F-Score   &Accuracy  \\
  	 \cline{1-5}
None	&95.82	&91.72	&93.73	&  \multirow{3}{*}{92.94}	\\
Intermediate		&86.81	&92.72	&89.67    &	        \\
Severe		&98.07	&99.47	&98.77    &	      \\
\hline
\multicolumn{5}{|c|}{ \textbf{RNN}} \\ 
\hline
\cline{1-5}
	\textbf{(n = 98113)} &Precision	&Recall	&F-Score   &Accuracy  \\
  	 \cline{1-5}
None	&95.56	&90.84	&93.14	&  \multirow{3}{*}{92.31}	\\
Intermediate		&85.50	&92.35	&88.79    &	        \\
Severe		&98.16	&99.33	&98.74    &	      \\
\hline
\end{tabular}

\end{center}

\end{table}  
\begin{table}[ht]
\caption{\label{tab:hyperall} All Model Hyper-Parameters. \textbf{Note:} MLP and RNN hyper-parameters only defer in Epoch, additionally, LMLP is an abbreviation for Linear MLP in this table}
\begin{center}       
\begin{tabular}{|l|c|l|c|} %% this creates two columns
%% |l|l| to left justify each column entry
%% |c|c| to center each column entry
%% use of \rule[]{}{} below opens up each row
\hline
\multicolumn{4}{|c|}{ \textbf{RNN Hyper-Parameters}} \\ 
\hline
\cline{1-4}
\textbf{Learning Rate}	&0.01		&\textbf{Hidden Layer Activation}	&None (LMLP),\\& & & ReLU (Others) \\
\hline
\textbf{Hidden Layers}		&[256, 256]	&\textbf{Batch Size}    &[8192, 8192]	        \\
\hline
\textbf{Epochs}		&80 (RNN), 50 (MLP)	
&\multirow{2}{*}{\textbf{Norm. Data}} &\multirow{2}{*}{Yes}	      \\

\cline{0-1}
\textbf{Optimizer}		&Adam & &\\
\hline
\end{tabular}

\end{center}

\end{table}  

To investigate glare prediction, ${g_{tp}}$, we remove the feature set ${\cal F}$ from the dataset, this removes all imagery components. What remains is what was extracted from image metadata. Without the feature set ${\cal F}$, 22 features remain, changing the input layer to 22 nodes ($\mathbb{R}^{22}$). Using metadata only on MLP and RNN results in what is shown in Table~\ref{tab:resMETA} using the same hyper-parameters shown in Table~\ref{tab:hyperall}.

\begin{table}[ht]
\caption{\label{tab:resMETA}MLP, and RNN Performance Predicting $g_{tp}$ }
\begin{center}       
\begin{tabular}{|l|c|c|c|c|} %% this creates two columns
%% |l|l| to left justify each column entry
%% |c|c| to center each column entry
%% use of \rule[]{}{} below opens up each row
\hline
\multicolumn{5}{|c|}{ \textbf{Metadata Only MLP}} \\ 
\hline
\cline{1-5}
	\textbf{(n = 98113)} &Precision	&Recall	&F-Score   &Accuracy  \\
  	 \cline{1-5}
None	&94.89	&88.75	&91.72	&  \multirow{3}{*}{89.22}	\\
Intermediate		&82.82	&88.59	&85.61    &	        \\
Severe		&83.61	&93.29	&88.18    &	      \\
\hline
\multicolumn{5}{|c|}{ \textbf{Metadata Only RNN}} \\ 
\hline
\cline{1-5}
	\textbf{(n = 98113)} &Precision	&Recall	&F-Score   &Accuracy  \\
  	 \cline{1-5}
None	&94.43	&88.13	&91.18	&  \multirow{3}{*}{88.71}	\\
Intermediate		&82.08	&88.12	&85.00    &	        \\
Severe		&83.59	&93.21	&88.14    &	      \\
\hline
\end{tabular}

\end{center}

\end{table}  
\section{Conclusion \& Future Works}
\label{sec:conc}
The generation of a large-scale dataset from sequential streams of imagery is necessary to advance the capabilities of this research. To suit this requirement a new dataset was generated using existing classifiers on newly acquired data, the CAM3 dataset. To label the CAM3 dataset an investigation into confidence tolerances and their effect on missing data was pursued. Note that this methodology assumes the FIT dataset is ergodic and places importance on limiting model labelling error based on the confidence tolerance, it makes decisions independent of missing error metrics. Choosing probability thresholds in this manner results in the labelling of 100 931 images from the CAM3 dataset. 15.83\% of the CAM3 dataset was rejected by the RF classifier. This is close to the predicted 26.13\% which might indicate two things: one, the CRFM classifier generalizes well to new data; two, the FIT dataset is ergodic. The 200 pseudo-labelled images randomly selected for manual evaluation yielded a 91\% accuracy which is within the 90\% confidence tolerance. Since land-based obstructions obscured many of the images analyzed during manual evaluation a high-resolution land-sea mask was employed to remove tainted imagery as best as possible. 

15.833\% of the original CAM3 dataset is made up of rejected pseudo-labels. It is expected that these rejected labels are partly made up of instances where certain glare conditions are present regardless of solar orientation. These aberrant conditions are of particular interest to our research goals, as they constitute a means of predicting and mitigating glare irrespective of solar orientation. Hence, they constitute a flight plan unencumbered by certain time of day constraints. 

A preliminary investigation into features that may impact pseudo-label rejection was performed, observing solar orientation normalized with respect to the camera. This small investigation yielded insufficient explanations, contrary to visual intuition. One reason for this might relate to the way the FIT dataset extracts aircraft bearing. To draw more concrete conclusions, we propose the following future work for pseudo-labelling. The FIT dataset's aircraft bearing feature will be re-analyzed using GPS tracks as opposed to the nearest available image which constitutes sparse instances of consecutive samples. This will be done to determine aircraft orientation more reliably. The re-analyzed FIT dataset will then be used once again in the same fashion presented in this manuscript. In order to gain a better understanding of pseudo-labelling bias, an investigation into the impact of all features on rejection will also be investigated.

The image formation model and the relationship between influencing features in glare forecasting and solar forecasting remain at the forefront of this research's guiding principles. Access to the pseudo-labelled CAM3 dataset allowed for an investigation into deep learning models such as MLP and RNN. They were investigated as a means of developing glare classification and prediction systems using features relating directly to imagery (used in classification only), solar orientation, and meteorological factors such as wind conditions, sea surface state, and cloud cover. 

Glare classification was first probed from a linear regressor constructed from a MLP without an activation function, trained and tested on the CAM3 dataset, to assess the complexity of the application. The result of this process yields an 82.76\% accuracy, which is similar to the accuracy of our three-class CRFM, 83.57\% trained and tested on the FIT dataset. Modest performance from linear regression might indicate that much of the CAM3 dataset is easily explainable. Adding a non-linear ReLU activation functions to MLP results in a significant boost to performance, 92.94\% accuracy. RNN sees slightly lower performance with an accuracy of 92.32\%. Glare prediction which does not use imagery to predict glare severity is examined and achieves promising results. Metadata only MLP and RNN attain accuracies of 89.22\% and 88.71\% respectively. It is observed that across all non-linear model results \emph{None} class precision remains high between 94.43\% and 95.82\%. This is likely due to pseudo-labelling bias from CRFM which is designed to perform best on \emph{None} class precision as it was identified that miss classifying a \emph{None} glare image would have the highest consequence on the detection function.  

Thoroughly understanding the factors influencing glare severity and how those factors impact the ability to spot megafauna is a central element to our ability to successfully estimate species populations and contribute to the longevity and conservation of critically endangered species. Predictive path planning based on an accurate glare model presents a reliable means of maximizing surface and subsurface visibility and increasing population estimation accuracy.

\subsubsection{Acknowledgements} The authors would like to thank Mylene Dufour, Marie-France Robichaud, and Maddison Proudfoot for their dedicated
work in labelling and preparing the FIT dataset used in our experiments. 
Stephanie Ratelle provided us with valuable insight on the challenges associated with megafauna surveys. The authors would also like to thank both Stephanie and Elizabeth Thompson for their crucial role in promptly granting us access to the CAM3 dataset and its associated flight tracks. 
Their experience in conducting systematic aerial
surveys throughout eastern Canadian waters was essential to
our work.

% latest springer style is alphabetically sorted.
\bibliographystyle{splncs04}
% this sorts by order of appearance in the paper
%\bibliographystyle{unsrt}
\bibliography{CVAUIbib}
\end{document}